\documentclass[conference]{IEEEtran}
\usepackage{float}
\usepackage{booktabs}
\usepackage{multirow}
\usepackage{algorithmic}
\usepackage{algorithm}

\IEEEoverridecommandlockouts
\usepackage{cite}
\usepackage{amsmath,amssymb,amsfonts}
\usepackage{algorithmic}
\usepackage{graphicx}
\usepackage{textcomp}
\usepackage{xcolor}
\usepackage{url}
\usepackage{hyperref}
\def\BibTeX{{\rm B\kern-.05em{\sc i\kern-.025em b}\kern-.08em
    T\kern-.1667em\lower.7ex\hbox{E}\kern-.125emX}}

\begin{document}
\DeclareRobustCommand*{\IEEEauthorrefmark}[1]{%
    \raisebox{0pt}[0pt][0pt]{\textsuperscript{\footnotesize\ensuremath{#1}}}}
\title{A Sensitivity-Driven Expert Allocation Method in LoRA-MoE for Efficient Fine-Tuning\\}

\author{
	\IEEEauthorblockN{
		Junzhou Xu\IEEEauthorrefmark{1,}\IEEEauthorrefmark{2}, 
		Boyu Diao\IEEEauthorrefmark{1,}\IEEEauthorrefmark{2}\IEEEauthorrefmark{*}, 
		Chengqiang Qi\IEEEauthorrefmark{1,}\IEEEauthorrefmark{2}, 
		Shaobo Zhao\IEEEauthorrefmark{1},
            Ruisheng Wang\IEEEauthorrefmark{1},
		Yongjun Xu\IEEEauthorrefmark{1,}\IEEEauthorrefmark{2}
        } 
	\IEEEauthorblockA{\IEEEauthorrefmark{1}Institute of Computing Technology, Chinese Academy of Sciences, Beijing, China}
	\IEEEauthorblockA{\IEEEauthorrefmark{2}University of Chinese Academy of Sciences, Beijing, China}
         \IEEEauthorblockA{\{xujunzhou23,qichengxiang23\}@mails.ucas.ac.cn} 
	\IEEEauthorblockA{\{diaoboyu2012,zhaoshaobo,wangruisheng,xyj\}@ict.ac.cn} 
    }

\maketitle

\begin{abstract}
As deep learning models expand, the pre-training-fine-tuning paradigm has become the standard approach for handling various downstream tasks. However, shared parameters can lead to diminished performance when dealing with complex datasets involving multiple tasks. While introducing Mixture-of-Experts (MoE) methods has alleviated this issue to some extent, it also significantly increases the number of parameters required for fine-tuning and training time, introducing greater parameter redundancy. To address these challenges, we propose a method for allocating expert numbers based on parameter sensitivity—LoRA-SMoE (A Sensitivity-Driven Expert Allocation Method in LoRA-MoE for Efficient Fine-Tuning). This method rapidly assesses the sensitivity of different tasks to parameters by sampling a small amount of data and using gradient information. It then adaptively allocates expert numbers within a given budget. The process maintains comparable memory consumption to LoRA (Low-Rank Adaptation) while ensuring an efficient and resource-friendly fine-tuning procedure. Experimental results demonstrate that compared to SOTA fine-tuning methods, our LoRA-SMoE approach can enhance model performance while reducing the number of trainable parameters. This significantly improves model performance in resource-constrained environments. Additionally, due to its efficient parameter sensitivity evaluation mechanism, LoRA-SMoE requires minimal computational overhead to optimize expert allocation, making it particularly suitable for scenarios with limited computational resources. All the code in this study will be made publicly available following the acceptance of the paper for publication.
Source code is at \href{https://github.com/EMLS-ICTCAS/LoRA-SMoE}{https://github.com/EMLS-ICTCAS/LoRA-SMoE} 

\end{abstract}


\section{Introduction}
In recent years, large pre-trained models have grown to encompass tens of billions or more parameters\cite{b15}\cite{b16}, rendering traditional full-parameter fine-tuning for specific downstream tasks extremely resource-intensive and time-consuming. To address this challenge, the research community has increasingly focused on Parameter-Efficient Fine-Tuning (PEFT)\cite{b17} strategies aimed at significantly reducing training costs while optimizing resource utilization efficiency.
To address this challenge, researchers have employed various strategies, including knowledge distillation\cite{yang2024clip}, distributed training\cite{dai2024sketch}, continual learning\cite{huang2024etag}\cite{liu2024continual}, prompt tuning \cite{ran2024brain}, and parameter-efficient fine-tuning, aiming to significantly reduce training costs while optimizing resource utilization efficiency.

One notable approach within PEFT is LoRA\cite{b18}, which primarily freezes the original model's parameters and introduces only a small number of additional parameters for training. These new parameters are typically embedded in key layers of the model through low-rank matrix decomposition. LoRA achieves rapid convergence while maintaining or even enhancing model performance, offering an attractive solution that drastically reduces computational complexity compared to traditional full-parameter fine-tuning methods. However, while fine-tuning can improve performance on downstream tasks, its effectiveness remains limited, especially in multi-task fine-tuning scenarios where shared parameters can lead to performance degradation.

Recent studies indicate that integrating low-rank adaptation with MoE models can effectively enhance performance in multi-task learning. 

MoLoRA\cite{b20} employs token-level soft routing mechanisms to achieve comparable results to full fine-tuning with less than 1$\%$ of trainable parameters. Meanwhile, LoRAMoE\cite{b19} introduces local balance constraint losses during the fine-tuning phase, mitigating catastrophic forgetting and significantly improving performance on downstream multi-task challenges. Moreover, MOLA\cite{b2} utilizes discrete routing policies—activating the top two experts with the highest routing weights each time—and adjusts the distribution of expert numbers across layers by increasing the number of experts in higher layers and decreasing them in lower layers. This approach enhances fine-tuning effects while maintaining the same volume of trainable parameters.

However, the allocation of expert numbers in the LoRA-MoE architecture still relies on manual settings, potentially leading to significant parameter redundancy and overfitting issues\cite{b14}, thereby weakening the model's generalization capability and downstream task performance. Inspired by "Sensitivity-Aware Visual Parameter-Efficient Fine-Tuning"\cite{b3}, the study reveals that different datasets exhibit varying sensitivities to the model's parameters, suggesting that tuning only the most sensitive parameters can substantially improve fine-tuning outcomes while reducing parameter redundancy. This approach overcomes the limitation of existing PEFT methods that rely on human-heuristic rules for introducing trainable parameters.

Based on these findings, we have designed a scheme that adaptively allocates the number of experts according to the sensitivity values of parameters across different tasks, under a fixed budget. We test three different sensitivity-based expert allocation methods. Experimental results show that our approach achieves better performance on eight common benchmark tests with a comparable number of trainable parameters to SOTA methods.

Our research further reveals that higher layers in the attention mechanisms of large models exhibit greater sensitivity to downstream tasks, with a few parameters in lower layers also showing sensitivity; in contrast, parameters in the middle layers of MLPs appear relatively insensitive. This discovery not only supports the existing conclusion that "Higher Layers Need More LoRA Experts"\cite{b2} but also delves into the differences in parameter importance across various downstream tasks. Our findings provide new insights and methodologies for building more efficient and flexible multi-task learning systems in the future.

\section{Background and Motivation}
\subsection{LoRA}
LoRA \cite{b18} focuses on reducing the number of parameters while keeping the original model's performance strong. It does this by freezing the weights of the pre-trained model and adding trainable low-rank matrices to each layer. This approach has proven to be an efficient fine-tuning method according to many test benchmarks. The forward computation for each linear layer works like this:
\begin{eqnarray}y & = & W_0x+BAx\end{eqnarray}
where $y$ is output and $x$ is input,$W_0$ is pre-trained model weight,$A \in \mathbb{R}^{d\times r}$ and $B \in \mathbb{R}^{r \times k}$ are both trainable low-rank matrices and $r\ll min(d,k)$.In the initial setup, matrix B is set to a zero matrix, and matrix A is initialized using the Kaiming Uniform initialization.
\subsection{MoE}
MoE is a neural network architecture that employs multiple small sub-models to process different types of input. Each expert focuses on specific tasks or data features, with inputs routed to the most suitable expert for processing, and the results are then aggregated. This approach allows the model to adaptively adjust its complexity based on the input, enhancing performance while maintaining computational efficiency.

LoRA-MoE\cite{b19} applies the LoRA method within the MoE framework, using low-rank matrices for fine-tuning within each expert. This enables each expert to flexibly adapt to specific tasks without significantly increasing parameters or computational cost. By doing so, LoRA-MoE combines the benefits of efficient fine-tuning and selective expert activation, providing stronger adaptability and performance improvements.

\subsection{Sensitivity}
Existing studies have shown that increasing the number of higher layers experts within the same budget can enhance model performance. However, this approach relies on manually set heuristic rules and lacks adaptability and efficiency. To address these limitations and explore a more efficient expert allocation mechanism, we were inspired by the sensitivity-aware visual parameter-efficient fine-tuning method\cite{b3}, which highlights the varying sensitivity of model parameters across different datasets. By fine-tuning only the most sensitive parameters, better adjustment effects can be achieved, moving away from the reliance on human-defined heuristics in traditional PEFT methods.

Based on this inspiration, we propose a new expert allocation strategy: assigning more experts to the parameters that are most sensitive according to task requirements. Specifically, we use the sum of squared gradients during backpropagation as an indicator of parameter sensitivity, with detailed calculations explained in Chapter Three.

To validate our strategy, we conducted a parameter sensitivity analysis on the Qwen2.5-3B-Instruct model using eight widely used datasets. The results showed that MLP layers exhibit higher parameter sensitivity values compared to attention layers, likely due to their distinct roles within the model architecture. We further illustrated this by plotting heatmaps, marking attention layers in red and MLP layers in blue, revealing common traits: higher-level attention layers are more sensitive to parameter changes, while MLP layers show higher sensitivity at lower levels and the top few layers. Notably, for almost all tasks, the middle layers of V and O matrices, as well as the VO matrix of the 0th layer and the down matrix of the first layer, exhibit significant parameter sensitivity.

Considering the similarities and differences among various tasks, our sensitivity-aware expert allocation method aims to optimize resource allocation by adaptively responding to these characteristics, thereby improving model performance. This method not only enhances the model's generalization ability but also provides valuable guidance for future research.
\begin{figure*}[!h]
  \centering
  \includegraphics[totalheight=7in]{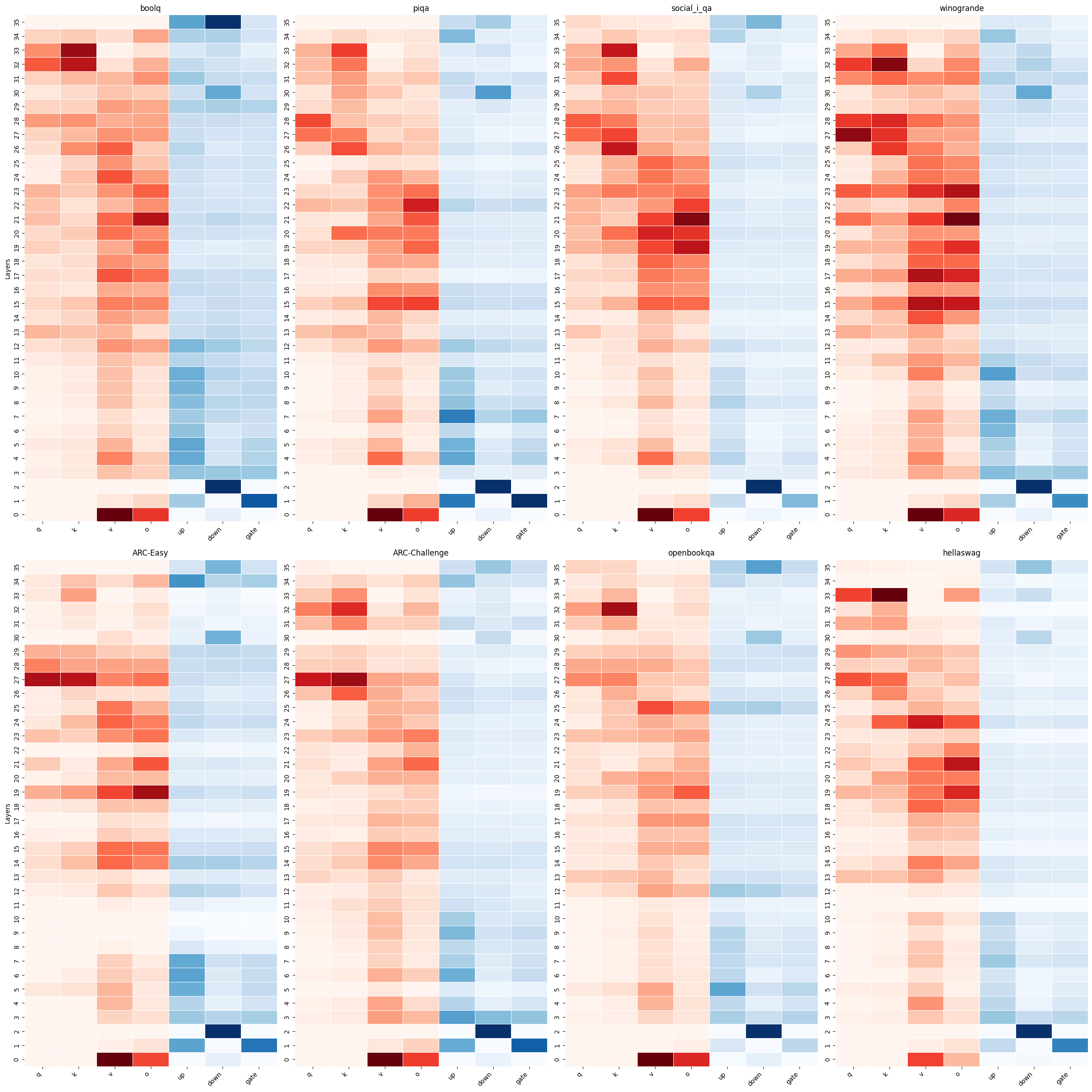}
  \caption{Parameter blocks sensitivity of Qwen2.5-3B-Instruct on eight datasets, the darker the color, the higher the sensitivity.} \label{fig:graph}
\end{figure*}
\subsection{Motivation}

Our work aims to reduce parameter redundancy in MoE models while striving to enhance model performance. Excessive parameters not only increase computational costs but can also lead to overfitting on training data. Therefore, we seek to eliminate experts that have minimal impact on model performance, retaining only those key parameters that significantly influence performance.

Based on the distribution of parameter sensitivity across different tasks within the model, higher-level parameters tend to be more sensitive for downstream tasks. This finding aligns with previous research perspectives. Thus, we adopt a method consistent with prior studies, allocating more parameter resources to higher-level attention layers within the same parameter budget as a baseline experiment. Additionally, to test our hypothesis, under an identical fixed budget, we focus on assigning more resources to high-sensitivity parameters. The primary distinction between these two approaches is that our method can capture low-level sensitive information and allocate more experts to a select few low-level sensitive parameters. Preliminary experimental results show that our method excels across eight test cases, maintaining robust performance even when further reducing the number of experts. This indicates that parameters with high sensitivity values are critical for task performance and should be allocated more experts to enhance their effectiveness.

Given the distributional differences among various matrix types, especially the overall higher sensitivity of MLP layers compared to attention layers, we propose three distinct strategies to address this imbalance. These strategies will be detailed in experimental sections, and validated through comparative analysis to assess their efficacy and applicability.
\section{LoRA-SMoE}
\begin{figure*}[!htbp]
  \centering
  \includegraphics[width=1.0\textwidth]{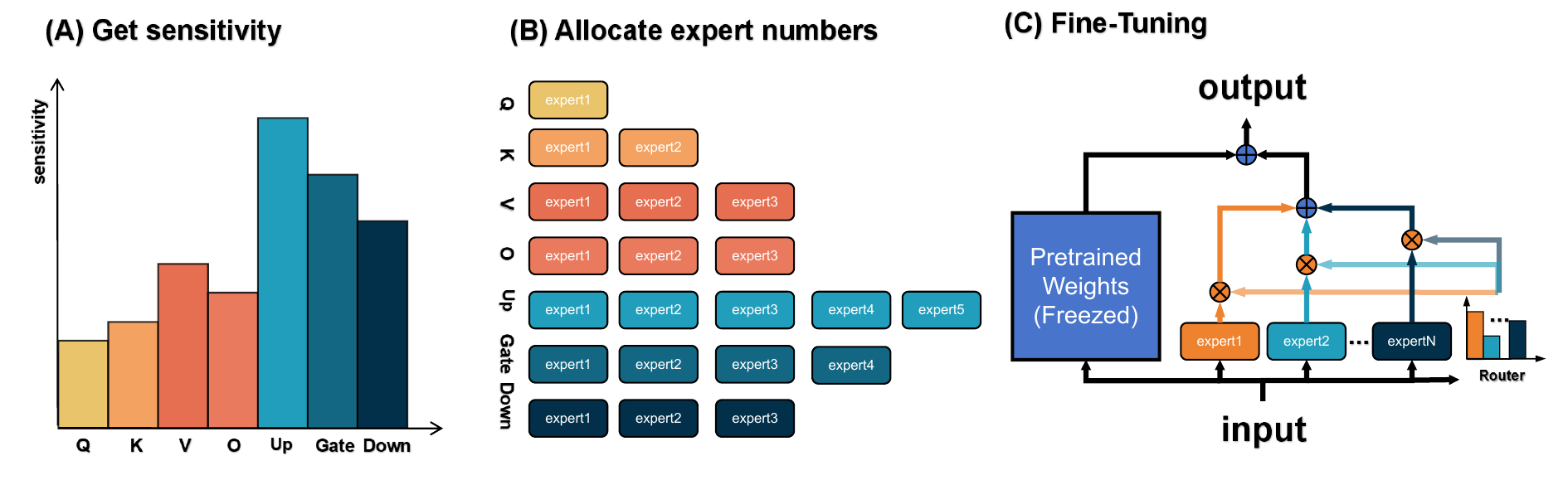} 
  \caption{LoRA-SMoE Framework Overview(steps A and B show our method for one layer's parameter blocks) } 
  \label{fig:示意图}
\end{figure*}
Fig \ref{fig:示意图} illustrates the steps of our method. First, we compute the sensitivity values for each parameter block. Next, we allocate the number of experts within a given budget. Finally, we apply the allocated expert numbers to the LoRA-MoE architecture for fine-tuning.
\subsection{Get Parameter Sensitivity}
To quantify the importance of model parameters across different tasks, we propose a method to compute parameter sensitivity. Let there be $T$ total tasks, each task $t$ has a training set $D_t$.We sample from each $D_t$ to obtain a smaller subset $C_t$, which is used to assess the sensitivity of various parts of the model's parameters.

Assuming the model has N parameter blocks, we divide these blocks into $M$ groups, where each group $m_i$ contains many parameter blocks, and all partitions collectively cover all parameter blocks.

Inspired by \cite{b3}, we define parameter sensitivity as the cumulative sum of the squares of gradients over multiple backpropagation iterations. This definition is based on that if a parameter has larger gradient changes over multiple forward and backward propagation iterations, then it plays a more important role in the task performance, thus indicating higher sensitivity. The specific algorithm steps are as follows algorithm \ref{alg:suanfa1}.
\begin{figure*}[!htbp]
  \centering
  \includegraphics[totalheight=3in]{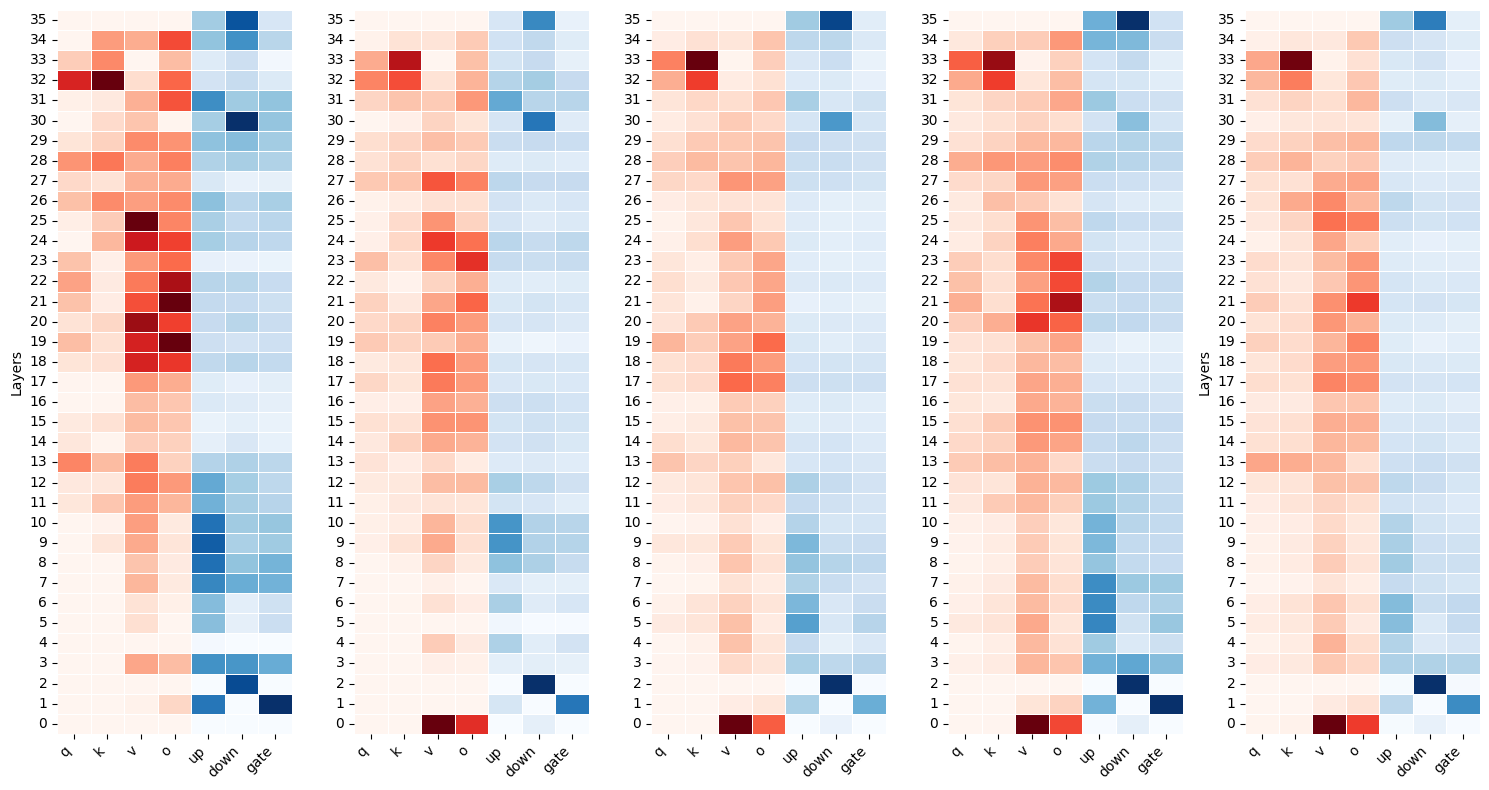}
  \caption{Parameter blocks sensitivity obtained with different sampling sizes} \label{fig:diffdatasets}
\end{figure*}

\begin{algorithm}[!h]
    \caption{Get parameter sensitivity}
    \label{alg:suanfa1}
    \renewcommand{\algorithmicrequire}{\textbf{Input:}}
    \renewcommand{\algorithmicensure}{\textbf{Output:}}
    \begin{algorithmic}[1]
        \REQUIRE model parameters $w$,samples set $C_t$  
        \ENSURE Sensitivity set $S=\{s_1,...s_N\}$    
        
        \STATE  Initialize $S=\{0\}^N$
         \FOR{each $m \in [1,M]$}
         
            \STATE   Freeze parameters $s_n \notin P_m$
            \STATE   Unfreeze parameters $s_n \in P_m$
            \FOR{each $i \in [1,C]$}
            
                \STATE Get $i$-th sample in $C_t$  
                \STATE Compute loss $E$
                \STATE Compute gradients $g$
                \FOR{each $n \in [1,N]$}
                    \STATE $s_n=s_n + g_n^2$
                \ENDFOR
            \ENDFOR
        \ENDFOR
    \end{algorithmic}
\end{algorithm}
\begin{table}
\caption{\textbf{means of parameter blocks's sensitivity under different sample sizes}}
\label{tab:t3}    
\centering
\setlength{\tabcolsep}{6pt} 
\begin{tabular}{c|cccccccccc}
\toprule
size& \multirow{1}{*}{Q}& \multirow{1}{*}{K}& \multirow{1}{*}{V}& \multirow{1}{*}{O}& \multirow{1}{*}{Up}& \multirow{1}{*}{Down}& \multirow{1}{*}{Gate}\\
\midrule
36&0.08&0.09&0.16&0.16&0.31&0.29&0.25&\\
72&0.13&0.16	&0.31&	0.29	&0.49	&0.54	&0.39\\
144	&0.24& 0.36 &0.55& 0.48 &0.97& 1.01 &0.71\\
288	&0.39 &0.54 &0.96 &0.87 &1.77 &1.74 &1.36\\
\bottomrule
\end{tabular}
\end{table}
\subsection{Adaptive Experts Allocation}
For most LLM models, including Qwen and Llama, the fine-tunable matrices include the QKVO (Query, Key, Value, Output) matrices of the attention layer and the up, down, and gate matrices of the MLP layer. To optimize these parameters for downstream tasks, we allocate experts based on the sensitivity values of each parameter block. Based on the uneven distribution of parameter blocks(see table \ref{tab:t3}), particularly with the MLP layers generally having higher values than the attention layers we propose three distinct methods:

\subsubsection{Unified Selection of Sensitive Parameters (LoRA-SMoE-U)}
Intuitively, parameters with higher sensitivity values have a greater impact on downstream task performance. Therefore, for each task within the budget, we select the most sensitive parameters from all parameter blocks and assign an expert to them. However, this approach may lead to an overallocation of experts to the MLP layer, as it typically exhibits higher sensitivity values.
\subsubsection{Separate Selection of Parameters from Self-Attention and MLP Layers (LoRA-SMoE-S)}
We independently select the number of experts within the budget for both the self-attention and MLP layers. This method allows for more precise control over resource allocation, ensuring that each module receives appropriate adjustments without disproportionately favoring any particular layer.
\subsubsection{Independent Allocation for Each Matrix (LoRA-SMoE-I)}
We designed a fine-grained approach where experts are independently allocated to each specific matrix, such as the Q, K, V, O matrices of the attention layer, and the up, down, and gate matrices of the MLP layer. This method minimizes the issue of uneven sensitivity value distributions across modules, ensuring optimal configuration for each component.
\subsection{Fine-tuning}
In our fine-tuning module, we employ the cutting-edge HydraLoRA\cite{b1}, featuring an asymmetric architecture. This design includes a central shared matrix A for learning shared task features and several B matrices for acquiring distinct features of individual tasks. By doing so, it reduces the number of model parameters while achieving better performance.
\begin{eqnarray}
y = W_0x+\sum_{i=1}^{E}\omega_iB_iAx
\end{eqnarray}
\begin{eqnarray}
w_i =softmax(Rx)
\end{eqnarray}
Here,$E$ represents the number of $B$ matrices, which is also the number of experts. A soft routing mechanism is employed to add the $B$ matrices together according to certain weights. The routing network is represented by a trainable dense layer of weights, which are processed using softmax to serve as the weights for each B matrix.
\section{Experiments}

\subsection{Tasks and Datasets}
LoRA-SMoE is designed to fine-tune multiple downstream tasks. To validate the effectiveness of our approach, we conducted evaluations across eight datasets: BoolQ\cite{b5}, PIQA\cite{b6}, SIQA\cite{b7}, Winogrande\cite{b9}, ARC-Easy\cite{b10}, ARC-Challenge\cite{b10}, OBQA\cite{b11}, and HellaSwag\cite{b8}.

BoolQ(Boolean Questions): This dataset contains questions and answers derived from real-world scenarios, including news articles and Wikipedia entries. Therefore, it requires a certain level of background knowledge to answer the questions accurately. It comprises 9,427 training instances and 3,270 test instances.

PIQA(Physical Interaction QA): Specifically crafted to assess AI systems' understanding of common-sense physics. It tests whether models can intuitively understand physical phenomena and apply common sense reasoning, requiring a degree of comprehension and inference to select the correct answer. The dataset includes 16,113 training samples and 1,838 test samples.

SIQA(Social Interaction QA): Designed to evaluate and research models' capabilities in social commonsense reasoning. The purpose of this project is to test whether models can understand human social behaviors, intentions, emotions, and causal relationships within social interactions. This dataset has 33,410 training examples and 1,954 test examples.

Winogrande: Used for evaluating models on pronoun resolution and complex semantic relationships. It consists of 63,238 training samples and 1,267 test samples.

ARC-E(AI2 Reasoning Challenge Easy): Targeted at assessing machines' ability to perform commonsense reasoning, with each question accompanied by four options, only one of which is correct. The dataset includes 2,251 training instances and 2,376 test instances.

ARC-C(AI2 Reasoning Challenge): Aims to evaluate and advance machines' capabilities in handling problems that require deep commonsense reasoning. It contains 1,119 training instances and 1,172 test instances.

OBQA(OpenBookQA): Evaluates machines on scientific knowledge and reasoning, created by the Allen Institute for AI. It encompasses 4,957 training instances and 500 test instances.

HellaSwag: Comprises multiple-choice questions, each providing a brief situational description followed by choices to select the most appropriate response. Questions cover a broad spectrum of everyday life scenarios, such as cooking, sports, and social interactions. It includes 39,905 training samples and 10,042 test samples.

\subsection{Experiment Settings}
We take Qwen2.5-3B-Instruct\cite{b12}, a 36-layer model, as our base model. To make the comparison fair, we adopted the following settings for all Experiments. We use AdamW\cite{b13} as the optimizer and the learning rate is 5e-5. We use Cosine learning rate decay\cite{b14}, and the decay rate is 1e-5. We set batch size, cutoff length, lora rank of each expert as 8, 512, 8. We set the number of samples C used to calculate parameter blocks sensitivities as 108 for all of our methods.
\begin{table*}
\caption{\textbf{Different tuning schemes's performance in eight tasks}}
\label{tab:t1}    
\centering
\begin{tabular}{l|cccccccccccc}
\toprule
Schemes& \multirow{1}{*}{\textbf{BoolQ}}& \multirow{1}{*}{\textbf{PIQA}}& \multirow{1}{*}{\textbf{SIQA}}& \multirow{1}{*}{\textbf{HellaSwag}}& \multirow{1}{*}{\textbf{WinoGrande}}& \multirow{1}{*}{\textbf{ARC-e}}& \multirow{1}{*}{\textbf{ARC-c}}& \multirow{1}{*}{\textbf{OBQA}}& \multirow{1}{*}{\textbf{average}}& \multirow{1}{*}{\textbf{Tuned/Total}}& \multirow{1}{*}{\textbf{time/hours}} \\
\midrule
base&49.1&79.7&71.7&73.7&61.8&92.1&78.2&73.6&72.5&$\-$ &$\-$\\
\toprule
LoRA(r=8)&68.9&84.6&78.3&89.6&71.1&93.4&84.3&86.2&82.0&0.48$\%$&0.48\\
LoRA(r=16)&68.0&84.1&77.5&89.8&71.7&94.6&84.9&86.2&82.1&0.97$\%$&0.50\\
LoRA(r=32)&69.1&84.8&78.8&89.8&72.8&94.4&84.1&86.8&82.6&1.90$\%$&0.57\\
MoLA-$\bigtriangledown$(8642)&69.2&84.9&79.3&90.1&73.7&93.9&85.3&86.0&82.8&1.38$\%$&1.35\\
HydraLoRA&69.9&84.9&78.8&89.6&74.4&93.9&85.2&86.4&82.9&2.51$\%$&1.31\\
\toprule
LoRA-SMoE-I(60$\%$)&69.5&85.0&79.7&89.9&73.1&94.4&84.2&87.0&82.85&1.50$\%$&1.16\\
LoRA-SMoE-U(60$\%$)&68.1&85.1&78.7&89.5&73.9&94.6&85.1&85.4&82.55&1.93$\%$&1.28\\
LoRA-SMoE-S(20$\%$)&67.0&84.9&77.6&88.8&71.5&93.9&83.7&85.2&81.3	&0.58$\%$&0.84\\
LoRA-SMoE-S(40$\%$)&67.6&84.8&78.5&89.5&72.6&94.1&84.5&86.4&83.0&1.12$\%$&1.08\\
LoRA-SMoE-S(60$\%$)&\textbf{70.1}&\textbf{85.2}&78.9&89.9&\textbf{74.5}&\textbf{94.8}&\textbf{85.6}&86.4&\textbf{83.2}&1.60$\%$&1.21\\
LoRA-SMoE-S(80$\%$)&68.8&84.3&\textbf{79.8}&\textbf{90.5}&73.2&93.8&85.0&\textbf{87.8}&82.9&2.04$\%$&1.25\\

\bottomrule
\end{tabular}
\end{table*}

\subsection{Preliminary Exploration of Parameter Redundancy}
To initially explore the degree of redundancy in model parameters, we conducted experiments on a dataset of size 15k under different budget conditions. Specifically, for each task, we allocated experts to the top 80$\%$, 60$\%$, 40$\%$, and 20$\%$ most sensitive parameter blocks. As a comparison, when this percentage reached 100$\%$, meaning every parameter block was assigned 8 experts, the method became equivalent to HydraLoRA\cite{b1}. We also fine-tuned different ranks of LoRA models.
The experimental results showed(see table \ref{tab:t1}) that as the ranks increased, the fine-tuning effect could be slightly improved but eventually reached a limit. We speculate that this might be due to interference between tasks caused by shared parameters. Our method achieved better results by retaining only the top 60$\%$  most sensitive experts, indicating that more experts do not necessarily lead to better performance and confirming the existence of expert redundancy. Therefore, in subsequent experiments, we consistently selected the top 60$\%$ most sensitive parameters for optimization.
\begin{table*}
\caption{\textbf{Expert selection consistency of different datasets and sampling sizes}}
\label{tab:t2}    
\centering
\setlength{\tabcolsep}{6pt} 
\begin{tabular}{c|cccccccccc}
\toprule
size& \multirow{1}{*}{BoolQ}& \multirow{1}{*}{PIQA}& \multirow{1}{*}{SIQA}& \multirow{1}{*}{HellaSwag}& \multirow{1}{*}{WinoGrande}& \multirow{1}{*}{ARC-e}& \multirow{1}{*}{ARC-c}& \multirow{1}{*}{OBQA}& \multirow{1}{*}{time}\\
\midrule
36&89.6&86.6&88.6&73.8&83.6&89.1&84.6&86.1&10s\\
72&91.5	&88.1	&86.5&	77.8	&86.6	&86.6	&88.1	&88.1	&18s\\
144	&94.1&89.1&89.6	&80.0&92.0&85.6&88.1&87.6&33s\\
288	&100&	100&	100&	100	&100&	100	&100	&100	&66s\\
\bottomrule
\end{tabular}
\end{table*}
\subsection{Method Comparison}
Furthermore, under the 60$\%$ setting, we compared the three proposed methods. For comparison, we also benchmarked against existing SOTA methods, including HydraLoRA\cite{b1} and MoLA-$\bigtriangledown $(8642)\cite{b2}. MoLA-$\bigtriangledown $(8642) is that for the 36-layer Qwen2.5-3B-Instruct model, the top 9 layers each allocate 8 experts, the next 9 layers each allocate 6 experts, and so on.

The experimental results (see table \ref{tab:t1}) show that our LoRA-SMoE-S achieved the best performance, LoRA-SMoE-I performed comparably to HydraLoRA with less number of parameters. In contrast, LoRA-SMoE-U exhibited relatively poorer performance. These findings indicate that allocating experts based only on parameter sensitivity is insufficient, it is also necessary to consider the differences between various modules.

Our research demonstrates that treating attention layers and MLP layers as independent entities and selecting the most sensitive parameter blocks within them for optimization yields better results. Specifically, in the case of LoRA-SMoE-S, we achieved superior performance compared to MoLA-$\bigtriangledown $ under the same parameters. This success is attributed to not simply increasing the number of experts in higher layers but rather flexibly allocating experts according to the sensitivity of parameters for each task, ensuring that lower-layer sensitive parameter blocks are also appropriately considered.

Additionally, our study confirms the presence of redundancy in LoRA-MoE models. To better capture task characteristics, it is crucial to allocate more experts to task-sensitive parameters. By avoiding unnecessary fine-tuning of insensitive parameters, our method effectively reduces the risk of overfitting and significantly enhances the model's generalization capability.

\subsection{Analyse sensitivity}
We conducted parameter blocks sensitivity analysis on the BoolQ dataset to understand how different sample sizes affect parameter blocks sensitivity. Fig \ref{fig:diffdatasets} shows the parameter blocks sensitivity distribution for sample sizes of 36, 72, 144, and 288. Table \ref{tab:t2} lists the expert-selected consistency results of the Qwen2.5-3B-Instruct model on different datasets. We compared the consistency between the results from different sample sizes and the largest sample size, using expert-selected outcomes at 60$\%$ redundancy. The sampling size is an integer multiple of the model parameter layers, as we update one layer of parameters at a step, thereby achieving a memory footprint comparable to that of LoRA fine-tuning. The results show that even with a very small sampling size, good consistency can still be achieved. Moreover, the time spent obtaining sensitivity values is very short compared to the entire fine-tuning process.

Our findings show that even with small sample sizes, the distribution of parameter sensitivity is very similar. On average, the sensitivity in MLP layers is higher than in attention layers, and there is high consistency between groups. This shows that our method is stable and efficient, maintaining consistent behavior regardless of the sample size used. It can be easily integrated into existing fine-tuning processes without adding a burden.

\section{Conclusion}
In this study, we propose an innovative parameter-efficient fine-tuning method that introduces parameter sensitivity analysis to achieve an adaptive allocation of the number of experts required for each parameter. We conducted comprehensive experimental evaluations on eight widely recognized benchmark datasets. The results show that this method not only significantly enhances the performance of LORA-MoE technology but also effectively addresses the issue of redundant expert resources. More importantly, our research reveals that, overall, higher layers in the attention mechanism exhibit higher sensitivity, whereas middle layers in the MLP demonstrate lower sensitivity. Additionally, we made some extra findings. These significant insights provide valuable guidance and new directions for optimizing the combination of MoE and PEFT, contributing to the advancement and technical progress in related fields.

\section*{Limitations}
Despite the encouraging results presented in this paper, certain limitations of our current study should be acknowledged. First, our analysis of sensitivity is primarily based on empirical observations rather than a rigorous theoretical explanation derived through mathematical derivation. Additionally, due to computational resource constraints, our experiments were conducted on a single Transformer model with a 3-billion-parameter scale, and we have not yet verified its effectiveness at larger parameter scales. Nevertheless, Our proposed method of adaptively allocating the number of experts based on parameter sensitivity is both simple and effective. Future research can build upon this foundation to further explore how to more efficiently leverage parameter sensitivity for optimal model tuning.

\section*{Acknowledgment}
This work is partially supported by the Chinese Academy of Sciences Project for Young Scientists in Basic Research (YSBR-107)

\bibliographystyle{IEEEtran}
\bibliography{IEEEabrv,sample}
\end{document}